\documentclass{svproc}
\usepackage{url}

\usepackage[subpreambles=false,mode=buildnew]{standalone}

\usepackage{import}

\usepackage[T1]{fontenc}
\usepackage[utf8]{inputenc}

\usepackage[english]{babel}

\usepackage{soul}

\usepackage{mathtools}

\usepackage{mathrsfs}
\usepackage{amsfonts}

\usepackage{amsmath}

\usepackage{subfig}

\usepackage{hhline}

\usepackage{array}

\usepackage{booktabs}

\usepackage[toc,page]{appendix}

\usepackage{pgfplots,tikz-3dplot}

\usepackage{colortbl}

\usepackage[ruled,vlined]{algorithm2e}

\DeclarePairedDelimiter\abs{\lvert}{\rvert}
\DeclarePairedDelimiter\norm{\lVert}{\rVert}
\DeclarePairedDelimiter\dprod{\langle}{\rangle}

\makeatletter
\let\oldabs\abs
\def\abs{\@ifstar{\oldabs}{\oldabs*}}
\let\oldnorm\norm
\def\norm{\@ifstar{\oldnorm}{\oldnorm*}}
\let\olddprod\dprod
\def\dprod{\@ifstar{\olddprod}{\olddprod*}}
\makeatother

\newcolumntype{C}[1]{>{\centering\arraybackslash}m{#1}}

\usetikzlibrary{decorations.text,calc,arrows.meta, backgrounds, intersections}
\tdplotsetmaincoords{70}{90}

\tikzset{image/.style={above right, inner sep=0pt, outer sep=0pt}}

\pgfplotsset{compat=1.15}
\usepgfplotslibrary{groupplots}

\pgfplotsset{select coords between index/.style 2 args={
                        x filter/.code={
                                        \ifnum\coordindex<#1\fi
                                        \ifnum\coordindex>#2\fi
                                }
                }}

\makeatletter
\def\relativepath{\import@path}
\makeatother

\usepackage[outdir=./epstopdf/]{epstopdf}

\begin{document}
\mainmatter              
\title{Hybrid Quadratic Programming - Pullback Bundle Dynamical Systems Control}
\titlerunning{Hybrid QP-Pullback Bundle DS}  
%
\author{Bernardo Fichera\inst{1} \and Aude Billard\inst{1}}
\authorrunning{Bernardo Fichera et al.} 
%
\tocauthor{Bernardo Fichera}
\institute{\'{E}cole polytechnique f\'{e}d\'{e}rale de Lausanne, Lausanne, Switzerland\\
\email{bernardo.fichera@epfl.ch}
}

\maketitle              

\begin{abstract}
  Dynamical System (DS)-based closed-loop control is a simple and effective way to generate reactive motion policies that well generalize to the robotic workspace, while retaining stability guarantees. Lately the formalism has been expanded in order to handle arbitrary geometry curved spaces, namely manifolds, beyond the standard flat Euclidean space. Despite the many different ways proposed to handle DS on manifolds, it is still unclear how to apply such structures on real robotic systems. In this preliminary study, we propose a way to combine modern optimal control techniques with a geometry-based formulation of DS. The advantage of such approach is two fold. First, it yields a torque-based control for compliant and adaptive motions; second, it generates dynamical systems consistent with the controlled system's dynamics. The salient point of the approach is that the complexity of designing a proper constrained-based optimal control problem, to ensure that dynamics move on a manifold while avoiding obstacles or self-collisions, is "outsourced" to the geometric DS. Constraints are implicitly embedded into the structure of the space in which the DS evolves. The optimal control, on the other hand, provides a torque-based control interface, and ensures dynamical consistency of the generated output. The whole can be achieved with minimal computational overhead since most of the computational complexity is delegated to the closed-form geometric DS.
\end{abstract}
%


\section{Introduction}
Reactive planning and control in the face of perturbation or sudden changes in the environment are key requirements in robotics. Dynamical System (DS) - based closed loop control has emerged as an effective technique to generate reactive policies. In a DS-based control, a policy is represented as a vector field $\mathbf{f} : \mathbb{R}^n \rightarrow  \mathbb{R}^n$ mapping a state-space variable of the robotic system $\mathbf{x} \in \mathbb{R}^n$ to an action in terms of velocity $\dot{\mathbf{x}} \in \mathbb{R}^n$, that the controlled system has to follow in order to achieve a certain task; i.e. $\dot{\mathbf{x}} = \mathbf{f}(\mathbf{x})$. The usage of DS in robotic control is advantageous, since it allows it to embed in a single closed-form law all the possible trajectories to accomplish a task. This yields instantaneous adaptation and recalculation of trajectories, providing robustness in face of both spatial and temporal perturbations.

Learning from Demonstration (LfD), a data-driven approach to learn DS bootstrapping from few observations of demonstrated tasks, has been in the last decades the main field of development of non-linear stable DS for closed-loop control. Starting from the Stable Estimator of Dynamical Systems (SEDS), the proposed approaches were gradually capable of learning more and more complex DS, retaining stability guarantees, either refining the stability constraints within an optimization problem, \cite{figueroa_physicallyconsistent_2018,ravichandar_learning_2017}, or adopting advanced diffeomorphism techniques, \cite{rana_euclideanizing_2020}. However all these approaches assume that the DS evolves along an Euclidean metric space.

Data encountered in robotics are characterized by varied geometries: For instance, joint angles lie on a d-dimensional torus ($T^d$) in revolving articulations, $3$-dimensional Special Euclidean groups ($SE(3)$) for rigid body motions, orientations represented as unit quaternions ($S^3$) and different forms of symmetric positive definite matrices such as inertia, stiffness or manipulability ellipsoids. Further, constraints may restrict the robot to a free sub-manifold which is difficult to represent explicitly. For example, an obstacle in the workspace can produce an additional topological hole that geodesics or “default” unforced trajectories should avoid.

Often such constraints are handled by constrained optimization. However, constrained optimization  scales poorly with the number and complexity of non convex constraints, making real-time Model Predictive Control (MPC) impractical. Sampling-based model predictive control, \cite{williams_model_2017,bhardwaj_storm_2021}, despite avoiding to solve explicitly an optimal control problem, heavily relies on the number and the length of trajectories, hence compromising accuracy for computation performances.

An alternative approach is to use Riemannian Motion Policies, a geometrical control framework designed to account for potentially non-Euclidean configuration and task spaces \cite{ratliff_riemannian_2018}. This approach not only has the advantage of handling arbitrary space geometry but can also account for the presence of the obstacles by locally deforming the space.

However, correctly designing metrics to account for such deformation directly in the configuration manifold can be difficult. The same applies for scenarios where multiple goals have to be taken into consideration. This motivated the usage of a tree structure where each node represents a sub-task connected to a primary task, \cite{cheng_rmpflow_2020}. This approach "split" the overarching goal in a series of sub-tasks, each described by a particular DS on a certain manifold. The overall DS representing the primary task is recovered via the so-called DS pullback. One of the main limitations of the this approach is the lack of geometric consistency. This translates in a control law that yields different output depending on the particular representation of a certain manifold structure.

To address this concern, \cite{bylard_composable_2021} introduced the Pullback Bundle Dynamical Systems (PBDS) framework for combining multiple geometric task behaviors into a single robot motion policy, while maintaining geometric consistency and stability. This approach provides a principle differential geometry description of the pullback operation for DS. It offers an easier and more intuitive way of designing geometrical entities, such as metrics on sub-task manifolds, and correctly captures the manifold structure due the geometric consistency property. Furthermore such approach proposes a clean and effective way to decouple the sub-task policy from the task priority metric design. Geometric formalization of such approach can be derived by considering more general Finsler geometries, \cite{ratliff_Generalized_2021}, \cite{vanwyk_Geometric_2022}. Yet one of the main limitations is the "absence" of the controlled system. While the overall PBDS is potentially capable of producing desired acceleration in configuration space, the designed DS law is agnostic of the robot model and generate infeasible dynamics.

Inverse dynamics (ID) torque control approaches based on Quadratic Programming (QP), \cite{feng_optimization_2014}, have gained increasing popularity by providing compliant motions and robustness to external perturbations. Similarly to classical optimal control approaches, ID with QP achieves accurate feasible trajectories by imposing constraints to satisfy model's dynamics. By considering acceleration and effort  as optimization variables, the optimal control problem can be formulated in a quadratic form. This offers high performance with guarantee of convergence to global optimum.

In this paper, we propose a hybrid Quadratic Programming Pullback Bundle Dynamical System yielding the following advantages: 1) torque-based control for compliant and adaptive motions; 2) model based approach for dynamical consistent motion. We propose a framework where much of the complexity of a classical QP problem can be avoided and "outsourced" to the geometrical DS framework. In this setting, constraints deriving from sub-manifolds motions or obstacle avoidance can be omitted into the formulation of the QP control problem because already embedded in the geometrical DS. On the other hand the limitations of geometrical DS such as the lack of velocity, acceleration, and control limits as well as dynamical constraints can be handled effectively by the QP control.


\section{Background}
\label{sec:background}
\vspace{-5mm}
Our notation follows \cite{carmo_riemannian_1992}. We employ the Einstein summation convention in which repeated indices are implicitly summed over. Given a set, $\mathcal{M}$, and a Hausdorff and second-countable topology, $\mathcal{O}$, a topological space $(\mathcal{M}, \mathcal{O})$ is called a \emph{d-dimensional} manifold if \( \forall p \in \mathcal{M} : \exists \mathcal{U} \in \mathcal{O} : \exists x : \mathcal{U} \rightarrow x(\mathcal{U}) \subseteq \mathbb{R}^d \), with $x$ and $x^{-1}$  continuous. $(\mathcal{U},x)$ is a \emph{chart} of the manifold $(\mathcal{M}, \mathcal{O})$. $x$ is called the \emph{chart map}; it maps $p \in \mathcal{M}$ to the point $x(p) = \left( x^1(p), \dots, x^d(p) \right)$ into the $\mathbb{R}^d$ Euclidean space. $\left( x^1(p), \dots, x^d(p) \right)$ are known as the coordinate maps or local coordinates. We refer to a point in $\mathbb{R}^d$ using the bold vector notation $\mathbf{x}$, dropping the explicit dependence on $p \in \mathcal{M}$. $x^i$ is the $i$-th local coordinate of $x \in \mathbb{R}^d$. We denote with $\mathcal{M}$ a \emph{differentiable Riemannian} manifold, that is a manifold endowed with a $(0,2)$ tensor field, $g$, with positive \emph{signature}. We refer to $g$ as a Riemannian \emph{metric}.

Let $\mathcal{M}$ be a Riemannian manifold. In local coordinates a second-order DS on $\mathcal{M}$ is expressed as
\begin{equation}
    \overbrace{\ddot{x}^k}^{\ddot{\mathbf{x}}} + \overbrace{\Gamma_{ij}^k \dot{x}^i}^{\Xi} \overbrace{\dot{x}^j}^{\dot{\mathbf{x}}}  = -\overbrace{g^{ak}}^{G^{-1}} \overbrace{\partial_a\phi}^{\nabla \phi} - \overbrace{D_m^k}^{D} \overbrace{\dot{x}^m}^{\dot{\mathbf{x}}},
    \label{eqn:full_ds}
\end{equation}
where $\Gamma_{ij}^k$ are the Christoffel symbols, $\phi$ a potential function on $\mathcal{M}$ and $D_m^k$ the components of the dissipative matrix; see \cite{bullo_geometric_2005}. Eq.~\ref{eqn:full_ds} can be expressed in the following vectorial form:
\begin{equation}
    \ddot{\mathbf{x}} = \mathbf{f}(\mathbf{x},\dot{\mathbf{x}}) = -G^{-1} \nabla \phi -D \dot{\mathbf{x}} - \Xi \dot{\mathbf{x}}.
    \label{eqn:vector_ds}
\end{equation}

Let $\mathcal{Q}$ and $\mathcal{X}$ be, respectively, a $m$-dimensional and a $n$-dimensional Riemannian manifolds, parametrized with respective local coordinates $\mathbf{q}$ and $\mathbf{x}$. We endow $\mathcal{X}$ with the Riemannian $h$; we use the capital notation $H$ to refer to the metric in matrix notation. The pullback of a DS onto $\mathcal{X}$ to $\mathcal{Q}$ reads as
\begin{equation}
    J \ddot{\mathbf{q}} = \overbrace{-H^{-1} \left( \nabla \phi + D J \dot{\mathbf{q}} \right) - \dot{J} \dot{\mathbf{q}} - \Xi J \dot{\mathbf{q}}}^{\mathbf{b}}.
    \label{eqn:coord_pullds}
\end{equation}
If the Jacobian, $J$, is injective everywhere, the solution of Eq.~\ref{eqn:coord_pullds} can be recovered via least mean square as an analytical solution to the optimization problem $ \min_{\ddot{\mathbf{q}}} \norm{ \mathbf{J} \ddot{\mathbf{q}} - \mathbf{b}}^2$.

\section{Method}
\label{sec:method}
\begin{figure}[ht]
    \vspace{-5mm}
    \centering
    \scalebox{.7}{\tikzstyle{block} = [draw, rectangle,
minimum height=3em, minimum width=6em]
\tikzstyle{sum} = [draw, circle, node distance=1cm]
\tikzstyle{input} = [coordinate]
\tikzstyle{output} = [coordinate]
\tikzstyle{pinstyle} = [pin edge={to-,thin,black}]

\begin{tikzpicture}[auto, node distance=2cm]
    \node [input, name=input] {};

    \node [block, right of=input, node distance=2cm] (dynamics) {$PBDS$};

    \node [block, right of=dynamics,
        node distance=3cm] (controller) {$QP$};

    \node [block, right of=controller,
        node distance=3cm] (system) {$SYSTEM$};

    \node [block, above of=controller,
        node distance=2cm] (model) {$MODEL$};

    \node [block, below of=controller] (kinematics) {$\mathcal{FK}$};

    \draw [->, transform canvas={yshift=3pt}] (dynamics) -- node[above,name=xd] {$\ddot{\mathbf{x}}$} (controller);
    \draw [->, transform canvas={yshift=-3pt}] (dynamics) -- node[below,name=qd] {$\ddot{\mathbf{q}}$} (controller);
    \draw [->] (controller) -- node[above,name=u] {$\tau$} (system);

    \node [right of=system, node distance=3cm] (output) {};



    \draw [->] (system) -- node [name=q] {$\mathbf{q},\ddot{\mathbf{q}}$} (output);
    \draw [->] (input) -- node[below, name=m] {} (dynamics);

    \node [input, above of=kinematics, name=n, node distance=1cm] {};

    \draw [-, transform canvas={xshift=-1.5pt}] (q) |- (n);
    \draw [->, transform canvas={xshift=-1.5pt}] (n) -| (dynamics);

    \draw [->, transform canvas={xshift=1.5pt}] (q) |- (kinematics);
    \draw [-] (kinematics) -| (input);
    \draw [-] (q) |- (model);
    \draw [->] (model) -- (controller);
\end{tikzpicture}}
    \caption{\footnotesize Block diagram of the control structure. (PBDS) the geometric DS generating the desired accelerations; (QP) Quadratic Programming controller; (SYSTEM) the actual controlled robotic system; (MODEL) dynamical model of the controlled robotic system; ($\mathcal{FK}$) the forward kinematics used to provide the task space to the geometric DS.}
    \label{fig:control_diagram}
    \vspace{-5mm}
\end{figure}
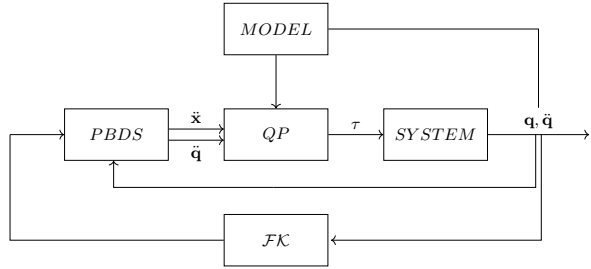

The control structure proposed is shown in Fig.\ref{fig:control_diagram}. Given as input both the configuration, $(\mathbf{q}, \dot{\mathbf{q}})$, and the "primary" task, $(\mathbf{q}, \dot{\mathbf{q}})$, space state, the geometric DS ("PBDS" block) generates both configuration and task space desired acceleration.
This information is processed by a QP controller ("QP" block) that, taking advantage of the dynamical model of the controlled system ("MODEL" block), yields the correct configuration space torques necessary to achieve the desired trajectory. The equations of motion and the constraint equations for an articulated robot system can be described as
\begin{equation}
    M(\mathbf{q})\ddot{\mathbf{q}} + h(\mathbf{q},\dot{\mathbf{q}}) = S\tau \\
    \label{eqn:robot_ds}
\end{equation}
where $M(\mathbf{q})$ is the inertia matrix, $ h(\mathbf{q},\dot{\mathbf{q}})$ is the sum of gravitational, centrifugal and Coriolis forces, $S$ is a selection matrix and $\tau$ is a vector of joint torques. Given a state, $(\mathbf{q}, \dot{\mathbf{q}})$, the equations of motion are linear in $\left[\ddot{\mathbf{q}} \quad \tau \right]^T$.

The Quadratic Programming problem has the following structure
\begin{equation}
    \min_{\mathbf{z}} \quad \frac{1}{2} \mathbf{z}^T \mathcal{W} \mathbf{z} + w^T \mathbf{z} \quad \text{s.t.} \quad C_E \mathbf{z} + c_E = 0, \quad C_I \mathbf{z} + c_I \ge 0.
    \label{eqn:qp}
\end{equation}
The unknown, $\mathbf{z}$, and constraints, $C_E$, $c_E$, $C_I$ and $c_I$, are problem specific. In our setting we have $\mathbf{z} = \left[\ddot{\mathbf{q}} \quad \tau  \quad \xi \right]^T$ as unknown variable. $\xi$ is a slack variable to relax the hard constraint imposed on the inverse dynamics as it will be clarified later. In order to define the cost functional we adopt the following matrices
\begin{equation}
    \mathcal{W} = \begin{bmatrix}
        Q & 0 & 0 \\
        0 & R & 0 \\
        0 & 0 & I \\
    \end{bmatrix},
    \qquad
    w^T = \begin{bmatrix}
        -\ddot{\mathbf{q}}_d^T Q & 0 & 0
    \end{bmatrix}.
\end{equation}
Given the matrices $\mathcal{W}$ and $w$ the cost in Eq.~\ref{eqn:qp} is quadratic form that tries to minimize the distance, $\ddot{\mathbf{q}} - \ddot{\mathbf{q}}_d$, between the current system acceleration and the desired acceleration provided by the geometric DS, while trying to minimize the effort, $\tau$, as well. We solve the inverse dynamics problem imposing it as a relaxed constraint with some slack variable $\xi$. The equality constraints matrices are
\begin{equation}
    C_E = \begin{bmatrix}
        M & -S & 0 \\
        J & 0  & I
    \end{bmatrix},
    \qquad
    c_E = \begin{bmatrix}
        h(\mathbf{q},\dot{\mathbf{q}}) \\
        \ddot{\mathbf{x}}_d - \dot{J}\dot{\mathbf{q}}
    \end{bmatrix}
\end{equation}
We use $C_I$ and $c_I$ to fullfil velocity, acceleration and torque constraints. No other "environment" related constrained is required since this part will be taken care from the geometric DS.

\begin{figure}[ht]
    \centering
    \subfloat[]{\scalebox{.6}{\subimport{../figures/}{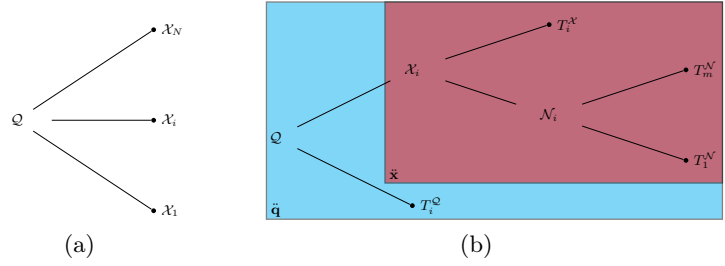}}
        \label{subfig:tree_simple}}
    \quad
    \subfloat[]{\scalebox{.6}{
\tikzstyle{level 1}=[level distance=3cm, sibling distance=3cm]
\tikzstyle{level 2}=[level distance=3cm, sibling distance=2cm]

\tikzstyle{bag} = [text width=4em, text centered]
\tikzstyle{end} = [circle, minimum width=3pt,fill, inner sep=0pt]

\begin{tikzpicture}[grow=right, sloped]
    \node (rect) at (4.8,0.6) [draw,thick,minimum width=10cm,minimum height=4.8cm, fill=cyan, opacity=.5, text opacity=1] {};
    \node (rect) at (6.1,1) [draw,thick,minimum width=7.4cm,minimum height=4cm, fill=red, opacity=.5, text opacity=1] {};
    \node at (0,-1.6) {$\ddot{\mathbf{q}}$};
    \node at (2.6,-0.8) {$\ddot{\mathbf{x}}$};
    \node[bag] {$\mathcal{Q}$}
    child {
    node[end, label=right:
    {$T^{\mathcal{Q}}_i$}] {}
    }
    child {
    node[bag] {$\mathcal{X}_i$}
    child {
    node[bag] {$\mathcal{N}_i$}
    child{
    node[end, label=right:
    {$T^{\mathcal{N}}_1$}] {}
    }
    child{
    node[end, label=right:
    {$T^{\mathcal{N}}_m$}] {}
    }
    }
    child {
    node[end, label=right:
    {$T^\mathcal{X}_i$}] {}
    }
    };
\end{tikzpicture}}
        \label{subfig:tree_diagram}}
    \caption{\footnotesize Tree diagram of the manifolds structure.}
    \label{fig:tree_diagram}
    \vspace{-5mm}
\end{figure}

The desired joint acceleration is produced by the "PBDS" block, see \cite{bylard_composable_2021} for details. Eq.\ref{eqn:coord_pullds} can be extended to an arbitrary number of task sub-manifolds, Fig.~\ref{subfig:tree_simple}. Let $f_i : \mathcal{Q} \rightarrow \mathcal{X}_i$ be $i$-th continuous map between the base space $\mathcal{Q}$ and the $i$-th target space $\mathcal{X}_i$.
On each of the target spaces $\mathcal{X}_i$ takes place a second order DS, of the type given in Eq.~\ref{eqn:vector_ds}. The DS parameters to be defined are: the mapping $f_i$, the metric of the target space $H$, the potential energy $\phi$ and dissipative coefficients $D$. Considering a weighted least mean square problem of the type $\min_{\ddot{\mathbf{x}}} \sum_i \norm{\mathbf{J}_i \ddot{\mathbf{x}} - \mathbf{b}_i}_{\mathbf{W}_i}^2$, where $\mathbf{W}_i \in \mathbb{R}^{s_i\times s_i}$ is task-weighting matrix with $s_i = \text{dim}\mathcal{N}_i$, we can derive an analytical solution for the second order dynamical systems on the base space $\mathcal{Q}$ as
\begin{equation}
    \ddot{\mathbf{q}} = \left( \sum_i J_i^T \mathbf{W}_i J_i \right)^{-1} \left( \sum_i J_i^T \mathbf{W}_i \mathbf{b}_i \right).
    \label{eqn:pbds}
\end{equation}
This process can be easily extended to arbitrary long "tree" of connected spaces where each target space may represent a base space for the following layer of spaces. The tree structure of the $PBDS$ block is shown in Fig.~\ref{subfig:tree_diagram}. It is clear that only on leaf nodes $T_(\cdot)^(\cdot)$ can be user-defined a second order DS. On the intermediate nodes, $\mathcal{X}_i$ or $\mathcal{N}_i$, as well as for the primary node, $\mathcal{Q}$, the flowing DS is automatically determined via the pullback operation.

We highlight that, even when combined with the QP problem in Eq.~\ref{eqn:qp}, the geometrical DS in Eq.~\ref{eqn:pbds} retains all the stability properties. The QP in Eq.~\ref{eqn:qp} solves (iteratively) an acceleration tracking problem generating the joint level torques that try to minimize the difference between the current joint acceleration and the desired acceleration produced by the geometrical DS. Further analysis and theoretical study should be dedicated to the convergence properties. Nevertheless when facing completely controllable/observable and co-located control problems, e.g. robotics arms or articulated systems, we believe that convergence does not represent a main issue, \cite{feng_3D_2013}, \cite{feng_optimization_2014}. Provided that the geometrical DS generates joint limits respecting trajectories we expect, at least for low frequency motion, good convergence properties.


\section{Preliminary Results}
\label{sec:results}
\begin{figure}[ht]
    \centering
    \subfloat[]{\includegraphics[scale=.1]{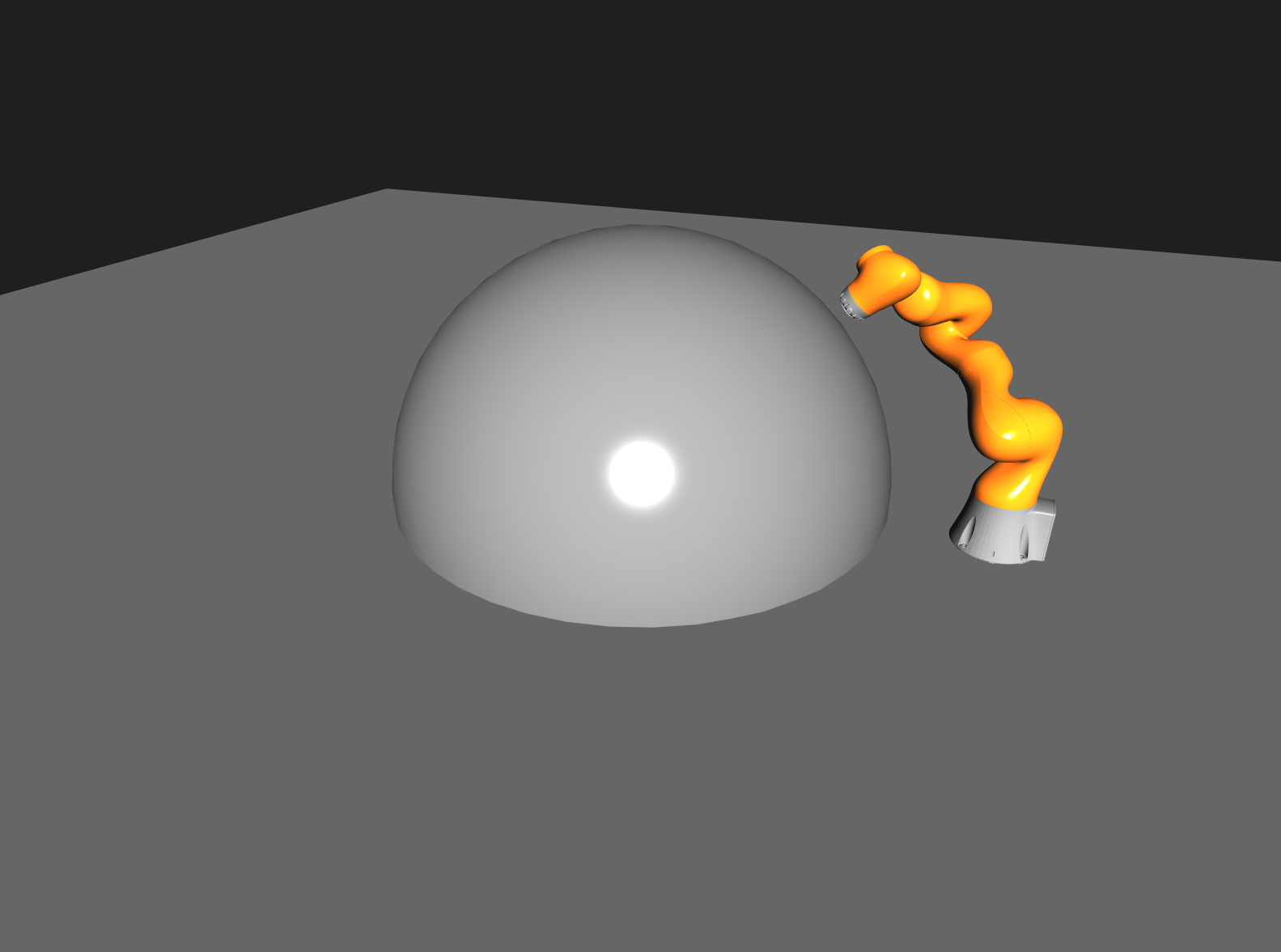}
        \label{subfig:sim_scenario}
    }
    \quad
    \subfloat[]{ \scalebox{.55}{
\tikzstyle{level 1}=[level distance=3cm, sibling distance=3cm]
\tikzstyle{level 2}=[level distance=3cm, sibling distance=2cm]

\tikzstyle{bag} = [text width=4em, text centered]
\tikzstyle{end} = [circle, minimum width=3pt,fill, inner sep=0pt]

\begin{tikzpicture}[grow=right, sloped]
    \node at (0,0.7) {($\ddot{\mathbf{q}}$)};
    \node at (2.9,0.7) {($\ddot{\mathbf{x}}$)};
    \node[bag] {$\mathcal{Q}$}
    child {
    node[bag] {$SE(3)$}
    child {
    node[bag] {$S^2$}
    child{
    node[end, label=right:
    {$\mathcal{O}_i$}] {}
    }
    child{
    node[end, label=right:
    {$\mathcal{P}$}] {}
    }
    child{
    node[end, label=right:
    {$\mathcal{D}$}] {}
    }
    }
    };
\end{tikzpicture}}
        \label{subfig:tree_scenario}
    }
    \caption{\footnotesize Simulated Environment: (a) the end-effector KUKA IWAA 14 follows a second order DS on the sphere while avoiding obstacles located on the sub-manifold; (b) manifolds tree to generate desired accelerations.}
    \label{fig:sim_env}
\end{figure}
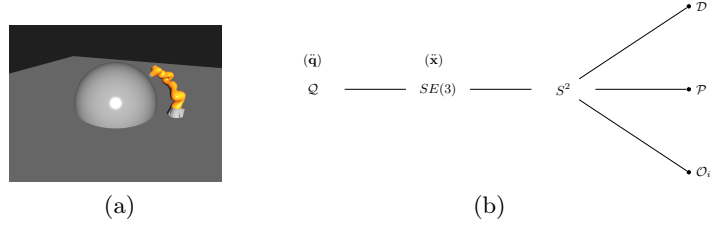

We tested the presented idea in scenario of constraint motion onto a sub-manifold in presence of obstacles. In Fig.\ref{subfig:sim_scenario} it is depicted the simulated environment.
The end-effector of a KUKA IWAA 14 has to follow a second order DS evolving on the sphere while avoiding obstacles along the path.

\begin{figure}[ht]
    \vspace{-3mm}
    \centering
    \subfloat[]{\includegraphics[scale=.15]{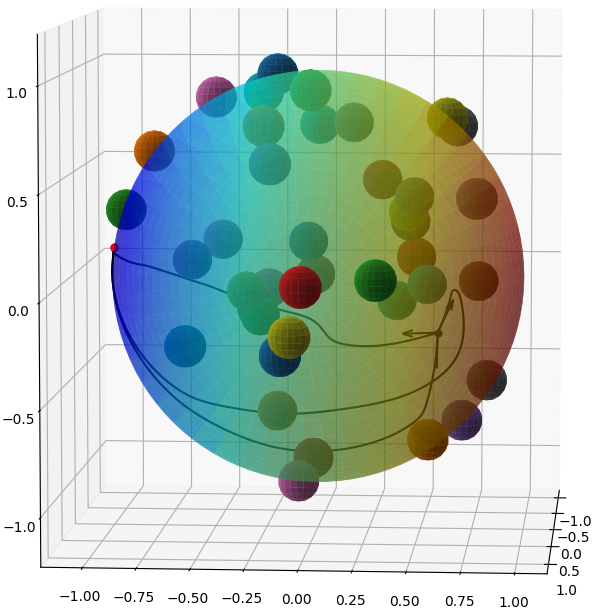}}
    \quad
    \subfloat[]{\includegraphics[width=.3\textwidth]{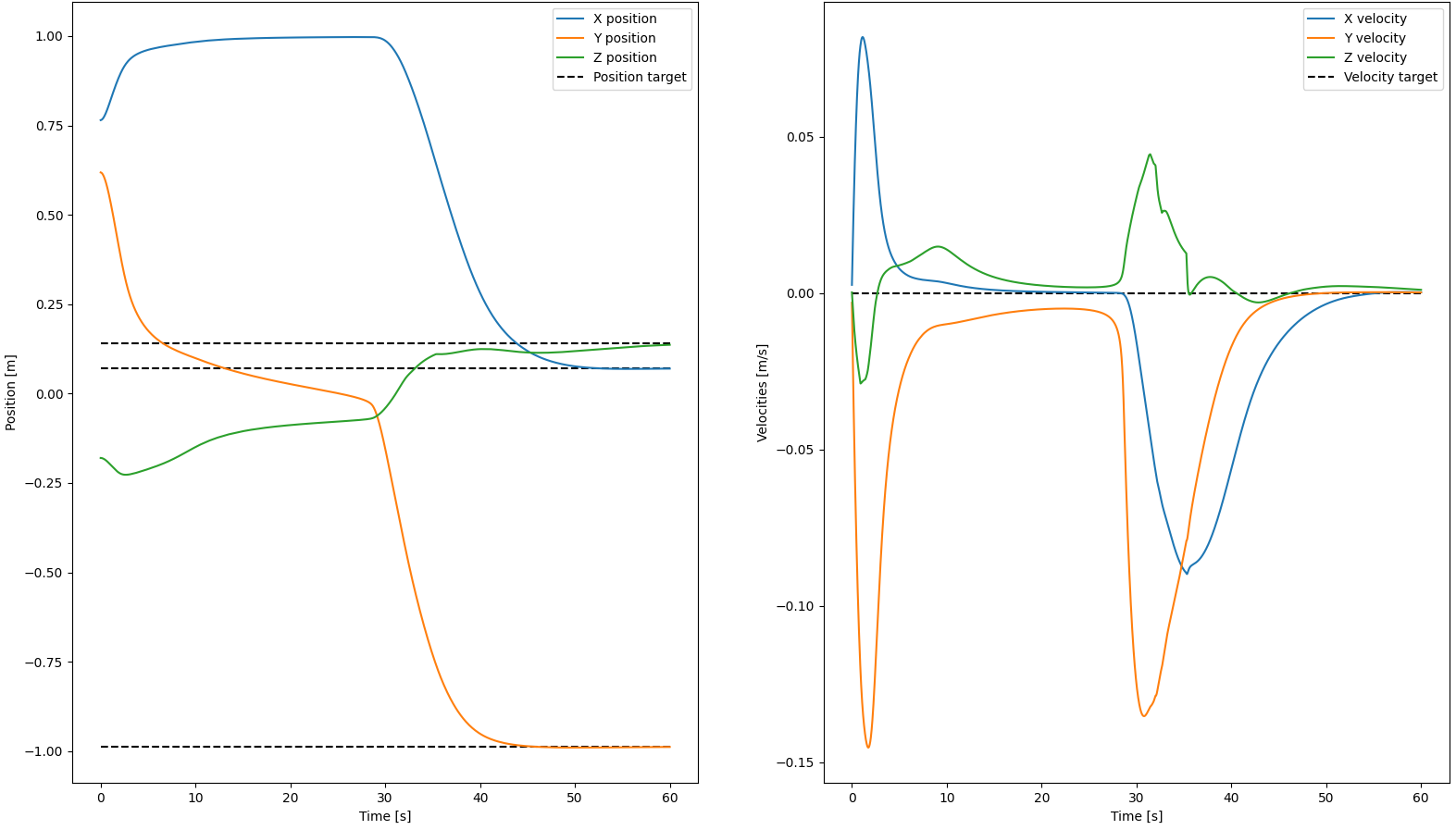}}
    \quad
    \subfloat[]{\includegraphics[width=.25\textwidth]{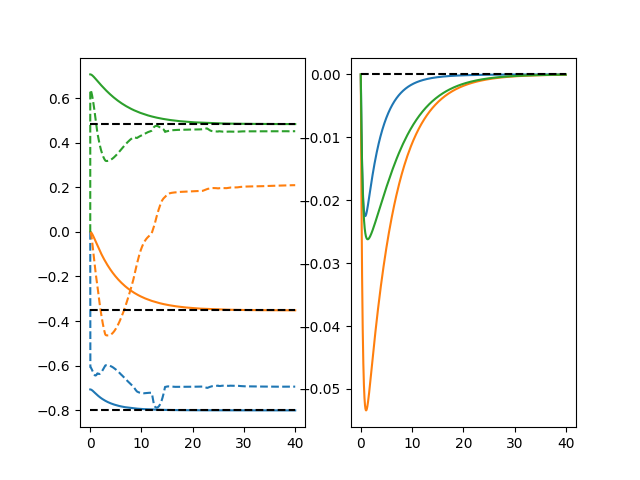}
        \label{subfig:results}}
    \caption{\footnotesize Pullback second order DS on $S^2$ (a) $3$ sampled trajectories; the color gradient represents the potential function used to generate the elastic force towards the desired attractor. (b) $x$, $y$, $z$ position and velocity signals. (c) End-effector trajectory tracking of the geometric DS.}
    \label{fig:sphere_motion}
    \vspace{-5mm}
\end{figure}

Fig.~\ref{subfig:tree_scenario} illustrated the structure of the manifolds tree designed in order to accomplish the desired task. At the top of the tree we have the configuration manifold, $\mathcal{Q}$, of controlled robotic system. It follows the Special Euclidean group, $SE(3)$. The map between $\mathcal{Q}$ and $SE(3)$ is the forward kinematics of the robot. Next we chose as sub-manifold for the constraint motion the $2$-dimensional sphere, $S^2$. The map between $SE(3)$ and $S^2$ performs a retraction onto the sphere for any point $x \in \mathbb{R}^3$ and $x \ni S^2$, while it project rotation matrices $R \in SO(3)$ such that the orientation of the end-effector remains perpendicular to the sphere.
\begin{table}[ht]
    \centering
    
\begin{tabular}{l|ccccc}
    \toprule
    Task Manifold                & Mapping                                                           & Metric          & Potential & Dissipation            & Weight Metric \\
    $\mathcal{P} = \mathbb{R}$   & $S^2 \rightarrow \mathbb{R} : p \mapsto \text{dist}(p, p_{goal})$ & $1$             & $||p||^2$ & $-d \dot{p}$         I & 1             \\
    $\mathcal{D} = S^2$          & $S^2 \rightarrow S^2 : p \mapsto p$                               & $g^{S^2}$       & 0         & 0                      & 1             \\
    $\mathcal{O}_i = \mathbb{R}$ & $ S^2 \rightarrow \mathbb{R} : p \mapsto \text{dist}(p, p_{obs})$ & $exp(a/(bx^b))$ & 0         & 0                      & 1             \\
    \bottomrule
\end{tabular}

    \caption{\footnotesize Parameters of the Sphere Bundle.}
    \label{tab:mapping_params}
    \vspace{-1cm}
\end{table}
For the sphere we define to two sub-manifolds tasks: one for the potential energy, $\mathcal{P}$, and one for the dissipative forces, $\mathcal{D}$. In addition arbitrary number of sub-manifolds tasks, $\mathcal{O}_i$, can be added in order to account for the presence of obstacles. Tab.~\ref{tab:mapping_params} the parameters for each sub-manifolds. Fig.~\ref{fig:sphere_motion} shows the pullback DS at level of $S^2$.
On the left, $3$ trajectories are sampled starting from different initial velocities. The convergence of the DS towards the attractor is shown on the center. Fig.\ref{subfig:results} shows the results about the quality of the end-effector trajectory tracking. In continuous line the it is reported the second order DS at level of $S^2$; the dashed shows the end-effector motion. For $x$ and $z$ axis the results shows an acceptable tracking error. $y$ axes on the other hand reports poor results due to the presence of joint limits constraints.
\vspace{-5mm}
\section{Conclusion}
\label{sec:conclusion}
\vspace{-3mm}
We introduced a structure to effectively perform closed-loop control into potentially non-Euclidean settings included but not limited to obstacle avoidance scenarios. Although very simple in its strategy, we believe that the introduce approach represents an effective way of taking into consideration within the control loop of potential dynamical constraints imposed by the robotic system at hand. The introduction of an iterative optimization process might be seen as a contradiction in DS framework that makes of the close-form approach a guarantee of reactivity. Nevertheless the Quadratic Programming nature of the low level optimization imposes a negligible sacrifice in terms of performance and it adds the capability of generating feasible trajectories.

%
%
\bibliographystyle{plain} 
\bibliography{references}

\end{document}